\documentclass[runningheads]{llncs}
\usepackage[T1]{fontenc}
\usepackage{graphicx}
\usepackage{comment}
\usepackage{orcidlink}
\usepackage{tikz}
\usepackage{subfigure}
\usepackage{marvosym}
\usepackage{multirow}
\usepackage{booktabs}
\usepackage{amsmath,amssymb,amsfonts}
\usepackage{diagbox}

\renewcommand{\orcidID}[1]{\orcidlink{#1}}
\begin{document}
\title{DiffS-NOCS: 3D Point Cloud Reconstruction through Coloring Sketches to NOCS Maps Using Diffusion Models}
\titlerunning{3D Reconstruction through Coloring Sketches to NOCS Maps Using DMs}
%
%
\author{Di Kong\inst{1,2,3}\orcidID{0009-0007-0771-5817} \and
Qianhui Wan\inst{4}$^{(\scalebox{0.85}{\textrm{\Letter}})}$}
\authorrunning{Di Kong and Qianhui Wan.}
%
\institute{Beijing University of Posts and Telecommunications, Beijing, China \\
\email{dikong@bupt.edu.cn} \and
Tsinghua University, Beijing, China \\
\and
Zhongguancun Academy, Beijing, China\\
\and
Beijing Normal University, Beijing, China \\ \email{wan\_qianhui@163.com}
}
\maketitle              
\begin{abstract}
Reconstructing a 3D point cloud from a given conditional sketch is challenging. Existing methods often work directly in 3D space, but domain variability and difficulty in reconstructing accurate 3D structures from 2D sketches remain significant obstacles. Moreover, ideal models should also accept prompts for control, in addition with the sparse sketch, posing challenges in multi-modal fusion. We propose DiffS-NOCS (Diffusion-based Sketch-to-NOCS Map), which leverages ControlNet with a modified multi-view decoder to generate NOCS maps with embedded 3D structure and position information in 2D space from sketches. The 3D point cloud is reconstructed by combining multiple NOCS maps from different views. To enhance sketch understanding, we integrate a viewpoint encoder for extracting viewpoint features. Additionally, we design a feature-level multi-view aggregation network as the denoising module, facilitating cross-view information exchange and improving 3D consistency in NOCS map generation. Experiments on ShapeNet demonstrate that DiffS-NOCS achieves controllable and fine-grained point cloud reconstruction aligned with sketches.

\end{abstract}
\section{Introduction}

Sketches, as abstract and efficient expressions, simplify complex ideas into simple lines and have long been used in 3D modeling. However, their geometric distortions and lack of texture, shading, and color make 3D reconstruction challenging, driving research to replicate this human capability in intelligent machines. Existing 3D generation and reconstruction methods \cite{zheng2023locally,kong2022diffusion,zhou20213d} achieve great stability by incorporating real 3D object data, benefiting from strong 3D priors. However, this comes at a high computational cost, particularly for high-resolution and fine-grained tasks, where training and inference times, as well as resource consumption, increase significantly. Meanwhile, 2D image-based approaches often rely on depth maps \cite{choe2021volumefusion} or normal maps \cite{wu2017marrnet}, which poorly represent the positional information of 3D point clouds and are primarily used as auxiliary supervision.

\begin{figure}[!t]\centering
	\includegraphics[width=8cm]{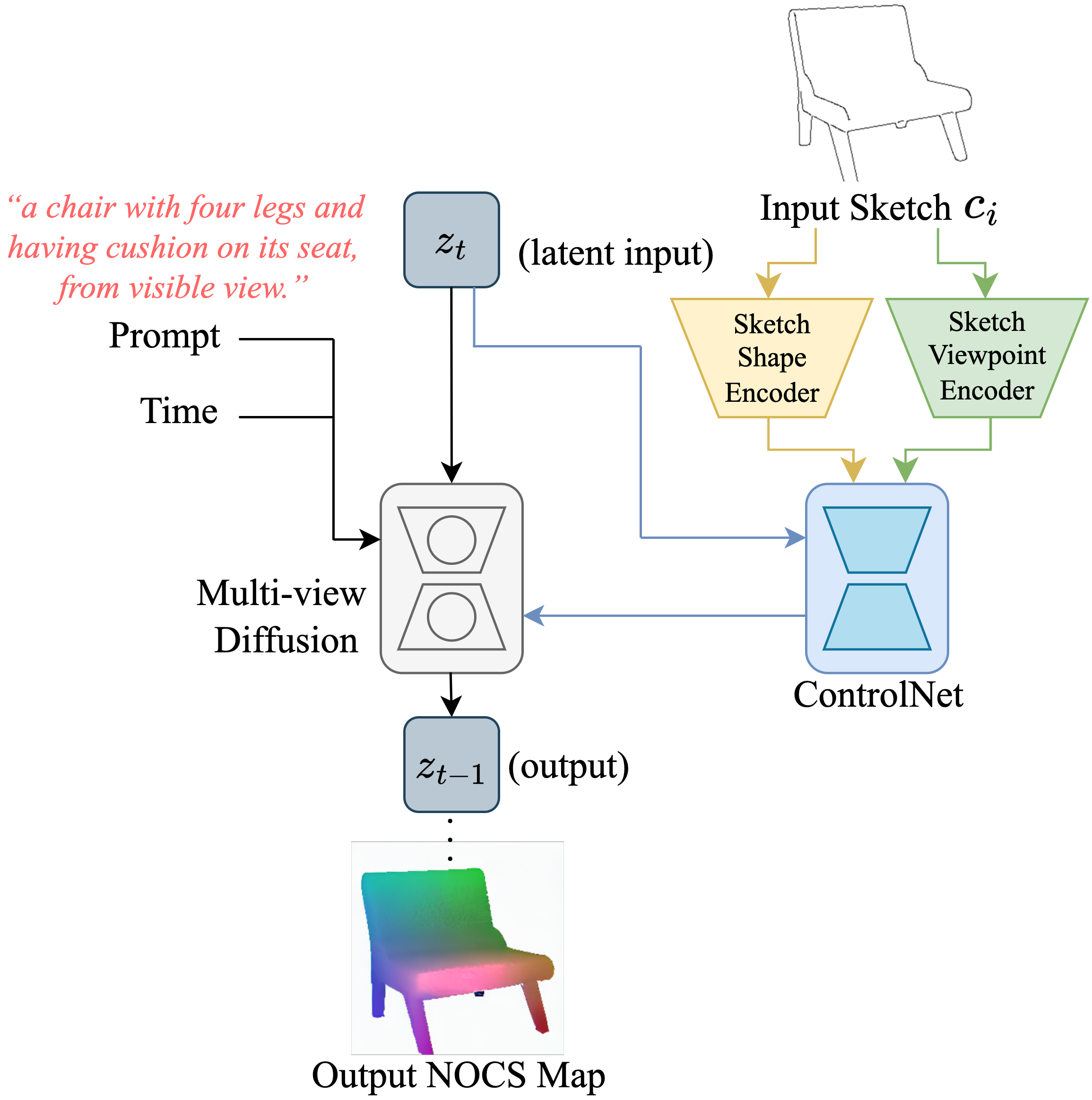}
	\caption{One-step denoising process for our DiffS-NOCS model.}
    \label{FIG_1}
\end{figure}

Can 3D objects be accurately reconstructed using only 2D training and inference, without real 3D priors? To bridge the gap between 2D images and 3D objects, we use Normalized Object Coordinate Space (NOCS) map \cite{wang2019normalized}, which predicts 3D point cloud positions directly from 2D images. A deep learning model extracts features from the 2D input and maps them to a NOCS representation for 3D reconstruction. In this work, we aim to generate NOCS maps from sketches, leveraging both front-view NOCS (Visible NOCS) and back-view NOCS (Occluded NOCS, X-NOCS) \cite{sridhar2019multiview}. The inclusion of X-NOCS addresses spatial ambiguity by incorporating shape and positional information from the backside of the object, enhancing reconstruction accuracy.

In recent years, diffusion models\cite{sohl2015deep,ho2020denoising,song2020score} have surpassed Generative Adversarial Networks (GANs)\cite{goodfellow2020generative,wang2021sketch} in many generative tasks \cite{saharia2022photorealistic,rombach2022high,zhang2023adding}, demonstrating exceptional ability to capture high-dimensional data structures in various dimensional spaces. In previous studies, the work of completing 2D image generation based on diffusion models is more mature than that of completing 3D generation. Inspired by their strong reasoning capabilities, we propose a conditional generative model for coloring sketches into NOCS maps using a modified ControlNet \cite{zhang2023adding}. However, generating corresponding NOCS maps from multi-view sketches poses two key challenges: (i) Sketch-based diffusion models often rely only on text prompts and shape features extracted from one input sketch, but our multi-view reconstruction task needs viewpoint information extracted from multiple sketches. (ii) Multi-view inputs share overlapping information across viewpoints. How to use this information and maintain consistency during the generation process is the key to multi-view reconstruction task. To address these, we train a viewpoint encoder to extract viewpoint information from sketches. This encoder is combined with the encoder of Stable Diffusion (SD) for contour extraction and feeds into a feature-level multi-view aggregation network, namely the multi-view decoder which is fine-tuned on the vanilla decoder of SD as a denoising network that facilitates cross-view information exchange and dynamically aggregates features to generate consistent outputs across varying input views. Such multi-view decoder, compared to the vanilla decoder in SD, is trained on multi-view NOCS maps and guarantees 3D consistency in the model's outputs, since more 3D priors are inherited.

\noindent \textbf{Our contributions.} We are the first to explore reconstructing 3D point clouds from multi-view sketches in 2D space using diffusion models. Our main contributions are: (i) We establish a novel connection between multi-view sketches and fine-grained 3D point cloud reconstruction via NOCS representation, enabling efficient 3D reconstruction in 2D space. (ii) We introduce a viewpoint encoder and a multi-view denoising network, ensuring controllable and consistent point cloud reconstruction for arbitrary viewpoint inputs. (iii) Experiments on synthetic and hand-drawn sketch datasets demonstrate that our model achieves excellent performance in 3D point cloud reconstruction from sketches.

\section{Related Works}
\subsection{Generative Diffusion Models}
Diffusion models \cite{ho2020denoising,sohl2015deep,song2020score} are a powerful class of generative models that have shown remarkable performance across diverse tasks, including high-resolution image generation \cite{dhariwal2021diffusion,ho2022cascaded}, biomedical image super-resolution \cite{kong2025d}, text-to-image generation \cite{nichol2021glide,saharia2022photorealistic,gu2022vector}, video generation \cite{wang2025lavie,gupta2024photorealistic} and 3D data generation \cite{zhou20213d,luo2021diffusion,kong2022diffusion,zheng2023locally}. Compared to GANs \cite{goodfellow2020generative} and Variational Autoencoders (VAEs) \cite{kingma2013auto}, diffusion models offer superior sample quality, stable training, and the ability to compute accurate likelihoods. Among these, Stable Diffusion \cite{rombach2022high} has gained prominence for its ability to generate high-quality images from text prompts. ControlNet \cite{zhang2023adding} extends Stable Diffusion by incorporating additional control inputs, allowing for more precise guidance during generation.

\subsection{Sketch-to-3D Generation and Reconstruction}
Research on sketch-based 3D generation and reconstruction can be categorized into single-view and multi-view approaches.

\noindent \textbf{Single-view.} A single-view sketch is drawn from a specific angle. Several approaches aim to generate 3D contents using single-view reconstruction. For example, Sketch2Mesh \cite{Guillard2021Sketch2MeshRA} leverages an encoder-decoder framework to create a latent representation of the input sketch, which is then refined by aligning the reconstructed 3D mesh's contours with the sketch during sampling. A diffusion model-based approach \cite{zheng2023locally}, LAS-Diffusion, uses 2D image block features to guide 3D voxel feature learning through a novel view-aware local attention mechanism, which greatly improves local controllability and generalization ability. Sketch2Model \cite{zhang2021sketch2model}, with single view sketch input but varying angles, explores the importance of viewpoint specification to overcome the reconstruction ambiguity problem and proposes a novel single view-aware generation method.

\noindent \textbf{Multi-view.} Multi-view sketches are drawn from multiple different angles. Theoretically, 3D reconstruction based on multi-view sketches can obtain a more accurate 3D model, but almost no one has conducted research in this direction. The 3D-R2N2 \cite{choy20163d} method is inspired by the progress of LSTM networks, along with single-view 3D reconstruction. And it proposes a new architecture that accepts one or more object instance images from different viewpoints and outputs 3D reconstruction results. It is important to note that 3D-R2N2 is not tailored for sketches, but for general images. Its performance decreases significantly when the input is sketch.

\section{Method}
In this section, we first briefly review the NOCS map, latent diffusion models and ControlNet. Then, we introduce the DiffS-NOCS model, highlighting the design of multi-modal content fusion and multi-view diffusion.

\subsection{Preliminary}
\begin{figure*}[!t]\centering
	\includegraphics[width=12cm]{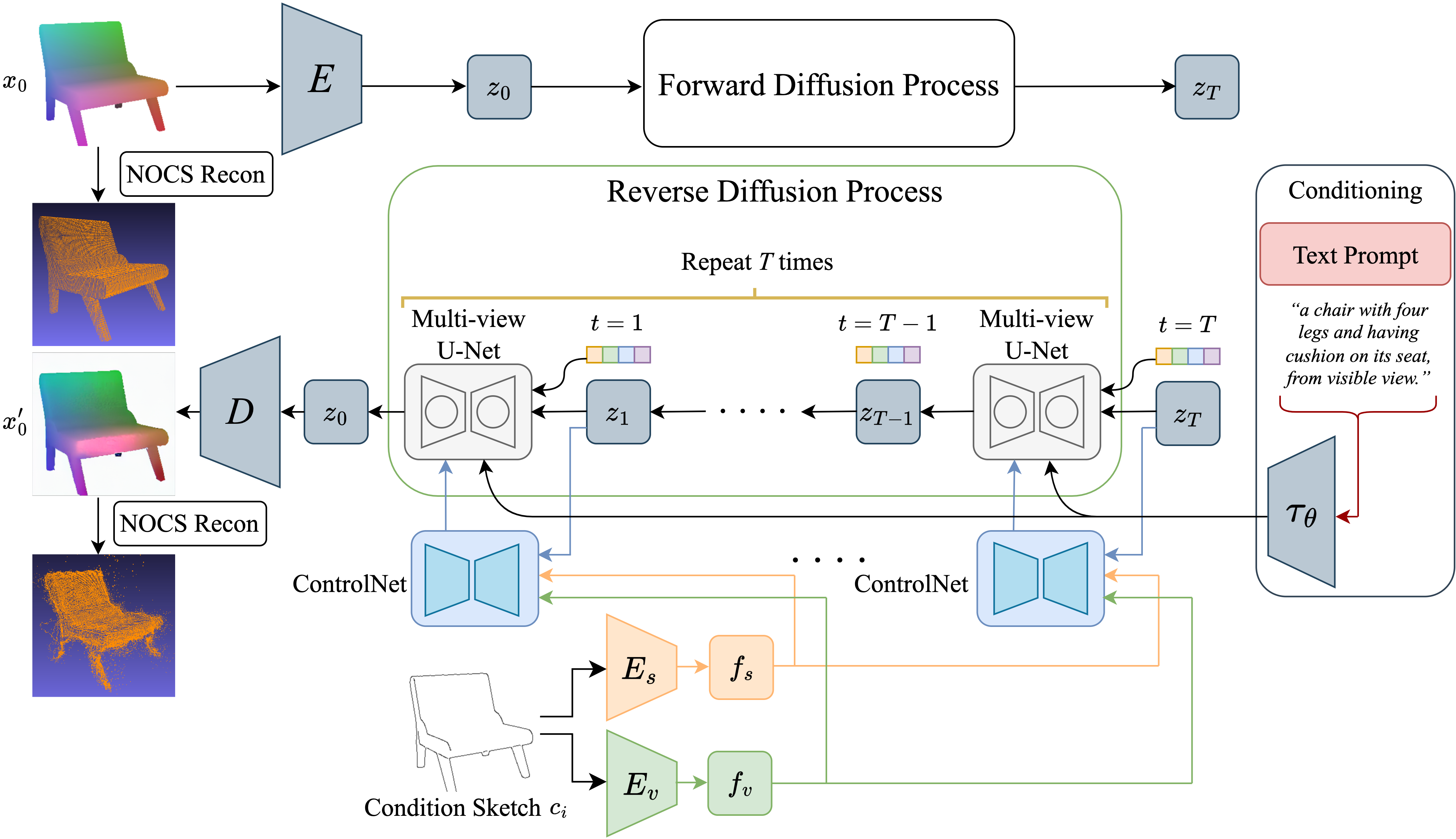}
	\caption{Overview of our proposed DiffS-NOCS model.}
    \label{FIG_2}
\end{figure*}

\noindent \textbf{NOCS Map.} We choose NOCS map as the 3D encoded representation, because it can capture the dense shapes of objects observed from any given viewpoint, while offering computationally efficient signal processing (such as 2D convolution). As shown in Figure \ref{FIG_2}, NOCS map is a two-dimensional projection of the instance three-dimensional NOCS point seen from a specific perspective. It contains 3 channels whose values are within $[0,1]$, representing the coordinates in the canonical space. In other words, each pixel in the NOCS map represents the three-dimensional position of the object point in NOCS (encoded by RGB). By reading out the pixel values, they can be easily converted to point clouds. The point clouds read out from different NOCS maps of the same object can also be directly combined together because they are all in the same canonical space.

\noindent \textbf{Latent Diffusion Models (LDM).} To deal with resources consuming problem in Denoising Diffusion Probabilistic Model \cite{ho2020denoising}, the LDM was proposed, which introduced a VAE to compress and encode the original object $x_0$ first, and then apply the encoded vector (the low-dimensional representation $z_0$ of the image) to the diffusion model. By adding the Attention mechanism to U-Net, the conditional variable $y$ is processed. The objective function of conditional LDM is:

\begin{align}
z_0 &= E(x_0), \nonumber \\
z_t &= \sqrt{\bar{\alpha}_t} z_0 + \sqrt{1 - \bar{\alpha}_t} \epsilon, \, \epsilon \sim \mathcal{N}(0, \mathbf{I}) \nonumber \\
\mathcal{L}_{\text{LDM}} &= \mathbb{E}_{t, z_0, \epsilon, y} \left[ \| \epsilon - \epsilon_\theta(z_t, t, \tau_\theta(y)) \|^2_{2} \right].
\label{equation_1}
\end{align}

Here, $y$ is the conditional data, and $\tau_\theta$ is responsible for processing $y$ into a feature vector. In this work, $y$ is a text prompt describing the object to be generated, and $\tau_\theta$ is the text encoder in the pre-trained CLIP model.

\noindent \textbf{ControlNet.} ControlNet is a neural network that controls a pre-trained latent diffusion model (eg. Stable Diffusion). It allows input of conditioned images, which are then used to guide image generation. In a pre-trained model, ControlNet incorporates an additional fully connected mapped condition $c$. This condition is introduced to the layer input, processed by a duplicate network mirroring the original layer’s structure, and then re-mapped before being added back to the layer's output.

\subsection{Multi-modal Content Fusion}
To reconstruct multi-view images, we first train a sketch viewpoint perception network to extract multi-view information from the input sketch. The network uses an encoder-decoder architecture based on a convolutional neural network (CNN) to predict the object's viewpoint category. We employ a pre-trained ResNet-50 \cite{he2016deep} (denoted $E_v$) as the encoder, and a simple fully connected network as the decoder to predict the viewpoint of the object. The encoder’s weights are initialized with ResNet-50’s pre-trained values and updated during training based on the sketch image and viewpoint label. The resulting Sketch Viewpoint Encoder encodes the input sketch $c_i$ into the viewpoint feature $f_v = E_{v}(c_i)$. This viewpoint feature $f_v$ is then concatenated with the shape feature $f_s = E_{s}(c_i)$ to form $c_f=F_{con}(f_v,f_s)$, where $F_{con}$ aggregates these two vectors. Finally, $c_f$ is passed to the NOCS map multi-view decoder $D_{MV}$ (in Sec. \ref{dmv}), providing it with both viewpoint and shape information for effective decoding. The addition of information and its placement are also crucial. While it may seem intuitive to integrate cross-modal information at the Encoder stage of U-Net for guiding the Decoder, we adopt a different approach. Instead of adding modules solely at key points, such as the Cross-Attention layer, we perform information fusion at the end of each layer, treating all added information equally.

\subsection{Multi-view Diffusion}
\label{dmv}
\begin{figure*}[!t]\centering
	\includegraphics[width=12cm]{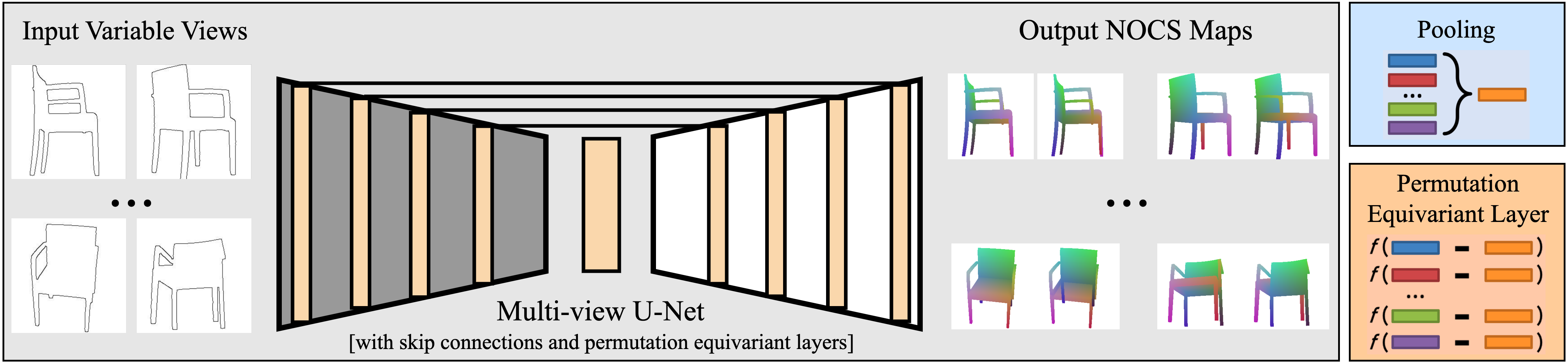}
	\caption{Feature level multi-view aggregation denoising U-Net.}
    \label{FIG_3}
\end{figure*}

We now move to the design of the multi-view diffusion model to conduct multi-view NOCS map prediction from input sketches. Multi-view images of objects contain significant overlapping information, and effectively utilizing this overlap is crucial for multi-view reconstruction. To enhance information exchange between different views and support varying numbers of input views, we design a feature-level multi-view aggregation denoising U-Net as the multi-view decoder $D_{MV}$ to replace the vanilla decoder in Stable Diffusion. Specifically, we propose using permutation equivariant layers that are independent of the view order.

The multi-view decoder $D_{MV}$ is illustrated in Figure \ref{FIG_3}. It mirrors the original decoder (vanilla $D_{LDM}$) except for the inclusion of several permutation equivariant layers (orange bars). A layer is considered envelope-equivariant if the diagonal elements of the learned weight matrix are equal to the diagonal elements. This is achieved by pooling the features from different views, subtracting the pooled features from each individual feature map, and then applying a nonlinear function. Namely, the permutation equivariant layer consists of mean subtraction followed by a convolutional nonlinearity. Multiple permutation equivariant layers (vertical orange bars) are applied after each downsampling and upsampling operation in the encoder-decoder of U-Net. Both average and maximum pooling can be used, but experimental results show that average pooling yields the best performance.

\noindent \textbf{Loss Function} The DiffS-NOCS model we propose follows the ControlNet design. We jointly train the multi-view decoder and ControlNet to generate NOCS maps from multi-view sketches. To incorporate viewpoint information, we replace the vanilla decoder in Stable Diffusion with a multi-view decoder, which includes permutation equivariant layers while retaining the pretrained weights from Stable Diffusion 2.1. The overall training objective consists of two components: (1) a denoising loss for multi-view decoder to learn an effective view-aware latent representation, and (2) a ControlNet-specific loss to ensure the generated NOCS maps align with the input sketches. The overall loss function is:
\begin{equation}
\mathcal{L} = \lambda_{\text{MV-LDM}} \mathcal{L}_{\text{MV-LDM}} + \mathcal{L}_{\text{ControlNet}}
\end{equation}
where $\mathcal{L}_{\text{MV-LDM}}$ is the updated LDM denoising loss defined in Equation \ref{equation_1}:
\begin{align}
\mathcal{L}_{\text{MV-LDM}} &= \mathbb{E}_{t, z_0, \epsilon, y, c_f} \left[ \| \epsilon - \epsilon_\theta(z_t, t, \tau_\theta(y), c_f) \|^2_{2} \right]
\end{align}
And the ControlNet loss consists of an L1 loss for pixel-wise supervision and a perceptual loss to preserve structural integrity:
\begin{equation}
\mathcal{L}_{\text{ControlNet}} = \lambda_1 \mathcal{L}_{L1} + \lambda_2 \mathcal{L}_{\text{perc}}
\end{equation}
where $\mathcal{L}_{L1} = \mathbb{E}_{\mathbf{x_0}, \hat{\mathbf{x_0}}} \left[ \|\mathbf{x_0} - \hat{\mathbf{x_0}} \|_1 \right]$ ensures direct alignment with the ground truth NOCS map $\mathbf{x_0}$, and $\mathcal{L}_{\text{perc}}$ minimizes the feature space difference using a pretrained VGG network:
\begin{equation}
\mathcal{L}_{\text{perc}} = \sum_i \|\phi_i(\mathbf{x_0}) - \phi_i(\hat{\mathbf{x_0}})\|_2^2
\end{equation}
where $\phi_i$ denotes the activation of the $i^{th}$ layer in the VGG network. We adopt a two-stage training strategy: (1) we first train ControlNet while freezing multi-view decoder to ensure effective conditioning, and (2) we jointly fine-tune both networks to optimize the overall generation quality. This approach ensures that ControlNet learns to modulate multi-view decoder effectively while allowing multi-view decoder to adapt to viewpoint-aware NOCS map reconstruction. $\lambda_{\text{MV-LDM}}$, $\lambda_1$ and $\lambda_2$ are the coefficients that balance each loss.

During training, we randomly replaced 50\% of the text prompts $y$ with empty strings to help DiffS-NOCS better interpret the input conditional sketch. This forces the model to rely more on the information of sketch, enhancing the DiffS-NOCS's ability to understand its semantic content. By leveraging the ControlNet architecture with the pre-trained Stable Diffusion model, the DiffS-NOCS model gains improved control over the NOCS map generation process, resulting in more accurate and well-aligned NOCS maps.

\section{Experiments}
\subsection{Experimental Setup}
\begin{figure}[!t]\centering
	\includegraphics[width=9cm]{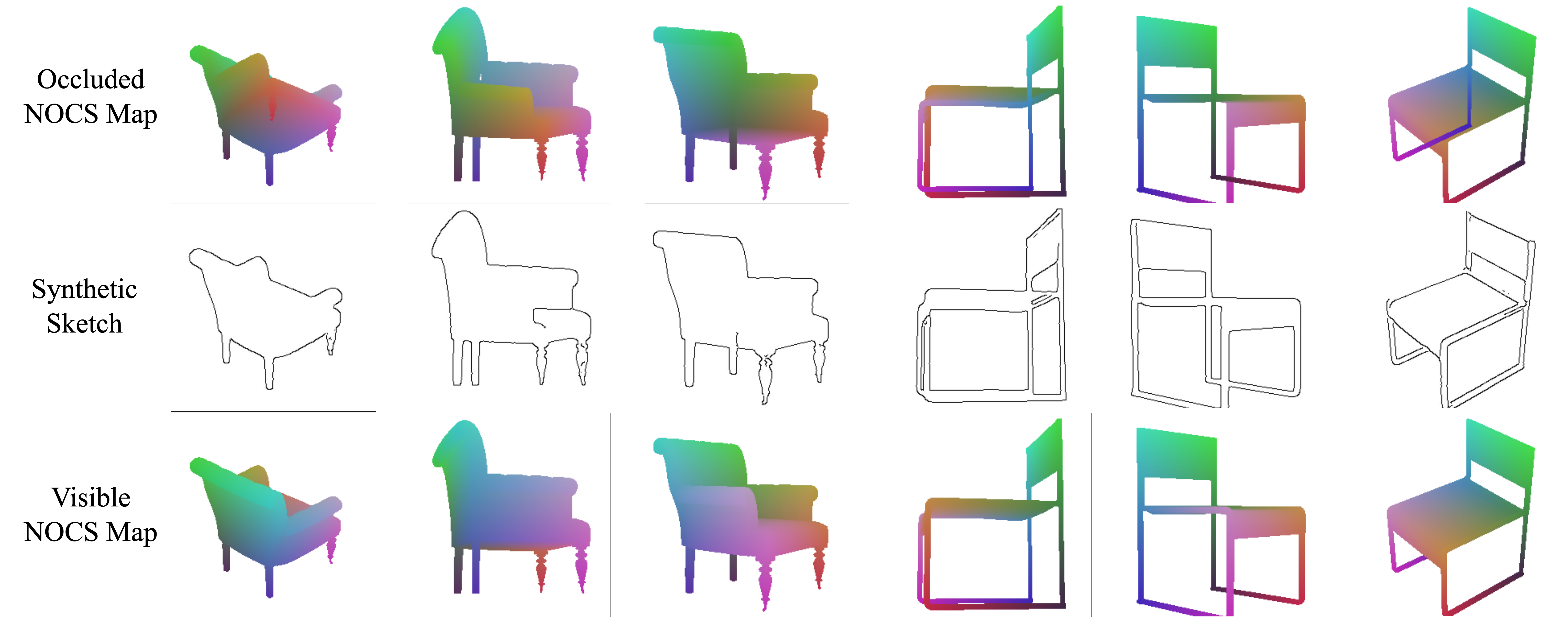}
	\caption{Composition of the \textit{S2N-ShapeNet} dataset.}
    \label{FIG_4}
\end{figure}

\noindent \textbf{Datasets}. To our knowledge, no dataset exists for sketch-NOCS map pairing. Therefore, we render the ShapeNet \cite{chang2015shapenet} dataset to generate NOCS map images that correspond to sketches, as shown in Figure \ref{FIG_4}. The resulting dataset, named \textit{S2N-ShapeNet}, includes three object categories: chair, car, and airplane, commonly used in related works. Tens of thousands of 3D assets from ShapeNetCoreV2 are utilized, with 20 views rendered for each object, five of which are randomly selected for training and testing. Text prompts are sourced from the Text2Shape \cite{chen2019text2shape}, which provides detailed descriptions for chairs. For cars and airplanes, we use the category names as prompts.

\begin{figure}[!t]\centering
	\includegraphics[width=9cm]{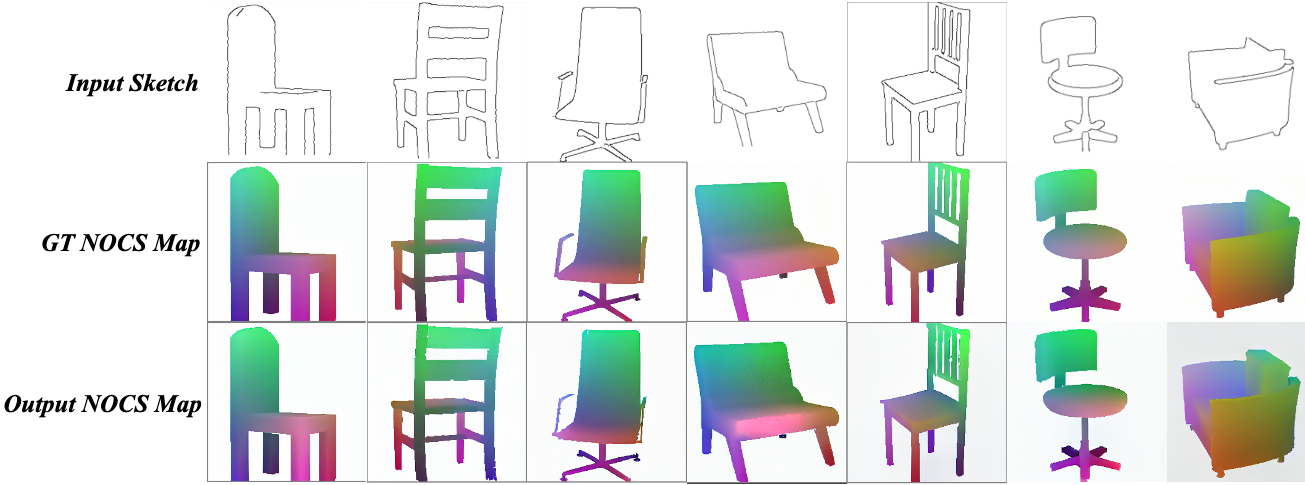}
	\caption{Results of Sketch-to-NOCS map generation.}
    \label{FIG_5}
\end{figure}

\noindent \textbf{Evaluation Metrics.} We use two standard evaluation metrics to assess the quality of the reconstructed 3D point cloud: Chamfer Distance (CD) and Earth Mover’s Distance (EMD) \cite{rubner2000earth}. Additionally, to ensure the generated 3D model closely matches the input sketch, we project the generated shape into the ground truth view and compute the 2D IoU score between the projected silhouette and the ground truth silhouette.

\noindent \textbf{Implementation Details.}
We train the sketch viewpoint perception network on two NVIDIA A100 40GB GPU cards with a batch size of 128, and the ControlNet-based DiffS-NOCS model with a batch size of 16, using Stable Diffusion 2.1 as the foundation model for multi-view diffusion model. In the training phase, the learning rate is set to $1 \times 10^{-5}$. And during inference, the DDIM steps for generating NOCS map images are set to 50.

\begin{table*}[!th]
  \centering
  \caption{Comparison of single-view and multi-view sketch-point cloud reconstruction performance.}
  \begin{tabular}{c|ccc|ccc|ccc}
    \toprule
    \multirow{2}*{\diagbox{Method}{Category}} &\multicolumn{3}{c}{CD $\downarrow$} &\multicolumn{3}{c}{EMD $\downarrow$} &\multicolumn{3}{c}{2D IoU $\uparrow$}\\
    & Chair & Car & Airplane & Chair & Car & Airplane & Chair & Car & Airplane\\
    \midrule
    LAS-Diffusion \cite{zheng2023locally} & 3.265  & 3.505 & 2.577 & 11.13 & 11.64 & 10.81 
               & 0.664  & 0.638 & 0.493     \\
    Sketch2Mesh \cite{Guillard2021Sketch2MeshRA} & 2.745  & 2.495 & N/A & 9.39 & 7.85 & N/A  
               & 0.726  & 0.840 & N/A     \\
    Sketch2Model \cite{zhang2021sketch2model} & 2.894  & 2.579 & 1.949 & 10.30 & 10.22 & 9.07  
               & 0.637  & 0.741 & 0.645     \\
    3D-R2N2 \cite{choy20163d} (1 view) & 3.079 & 3.129 & 2.147 & 9.75 & 8.96 & 6.94  
               & 0.618 & 0.697 & 0.566      \\
    Ours (1 view)     & \textbf{2.304} & \textbf{2.316} & \textbf{1.318} & \textbf{8.96} & \textbf{7.18} & \textbf{6.34} 
                & \textbf{0.832} & \textbf{0.921} & \textbf{0.779}      \\
    \midrule
    3D-R2N2 (2 views)   & 2.963  & 2.923 & 1.798 & 9.52 & 7.83 & 6.90  
           & 0.723  & 0.814 & 0.613     \\
    3D-R2N2 (3 views)   & 2.649  & 2.871 & 1.460 & 9.27 & 7.52 & 6.92  
           & 0.745  & 0.839 & 0.642     \\
    3D-R2N2 (5 views)   & 2.508  & 2.450 & 1.239 & 9.19 & 7.31 & 6.52  
           & 0.764  & 0.887 & 0.674     \\
    Ours (2 views)   & \textbf{1.915}  & \textbf{1.958} & \textbf{1.080} & \textbf{8.82} & \textbf{7.07} & \textbf{6.28}  
           & \textbf{0.836}  & \textbf{0.938} & \textbf{0.791}     \\
    Ours (3 views)   & \textbf{1.657}  & \textbf{1.644} & \textbf{1.149} & \textbf{8.31} & \textbf{6.82} & \textbf{6.23}  
           & \textbf{0.848} & \textbf{0.954} & \textbf{0.827}     \\
    Ours (5 views)   & \textbf{1.324}  & \textbf{1.732} & \textbf{0.674} & \textbf{8.05} & \textbf{6.59} & \textbf{6.15}  
           & \textbf{0.875} & \textbf{0.972} & \textbf{0.854}     \\
    \bottomrule
  \end{tabular}
  \label{table1}
\end{table*}

\begin{table}[t!]
  \centering
  \caption{Quantitative evaluations on the ProSketch dataset.}
  \begin{tabular}{m{2.8cm}<{\centering}|m{1.5cm}<{\centering} m{1.5cm}<{\centering} m{1.5cm}<{\centering}}
    \hline
    Method & {CD $\downarrow$} & {EMD $\downarrow$} &{2D IoU $\uparrow$} \\
    \hline
    LAS-Diffusion \cite{zheng2023locally} & 2.936 & 10.07 & 0.698 \\
    Sketch2Mesh \cite{Guillard2021Sketch2MeshRA} & 6.749 & 12.37 & 0.604 \\
    Sketch2Model \cite{zhang2021sketch2model} & 5.825 & 11.72 & 0.557 \\
    Ours & \textbf{2.853} & \textbf{9.95} & \textbf{0.728} \\
    \hline
  \end{tabular}
  \label{table_prosketch}
\end{table}

\subsection{Evaluation Results}
\noindent \textbf{Quantitative Comparison.}
(1) \textbf{Single-View Reconstruction.} For mesh-represented methods, meshes are converted to point clouds, sampled to match our output size, and scaled to a unit diagonal bounding box. For 3D-R2N2, surface voxels are extracted, converted to points, and scaled accordingly. Table \ref{table1} shows that our method outperform all baselines in all categories on synthetic sketch dataset. We also test the generalization ability of our model on a hand-drawn sketch dataset ProSketch-3DChair \cite{zhong2020towards} without fine-tuning. The statistics results in Table \ref{table_prosketch} demonstrate our method are much more faithful to the input sketch. (2) \textbf{Multi-View Reconstruction.} We compare our model with 3D-R2N2. Both methods are trained on random 5-view sketch-NOCS map pairs per batch and evaluated with up to 5 views at inference. Table \ref{table1} shows that our method outperforms 3D-R2N2 in all metrics, demonstrating better geometric consistency and shape matching.

\begin{figure}[!t]\centering
	\includegraphics[width=12cm]{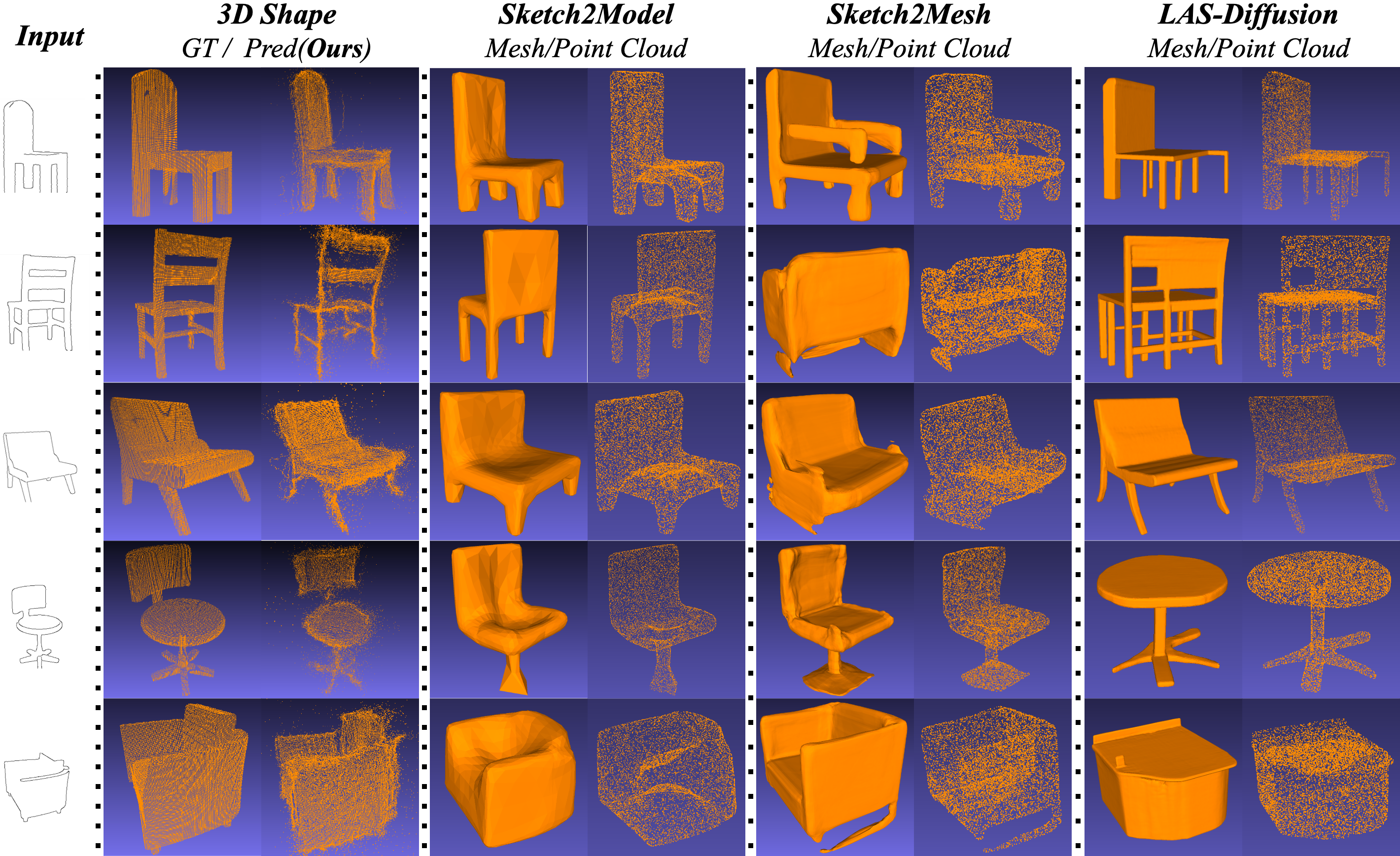}
	\caption{Qualitative comparison of different models on synthetic chair sketches.}
    \label{FIG_6}
\end{figure}

\begin{figure}[!t]\centering
	\includegraphics[width=12cm]{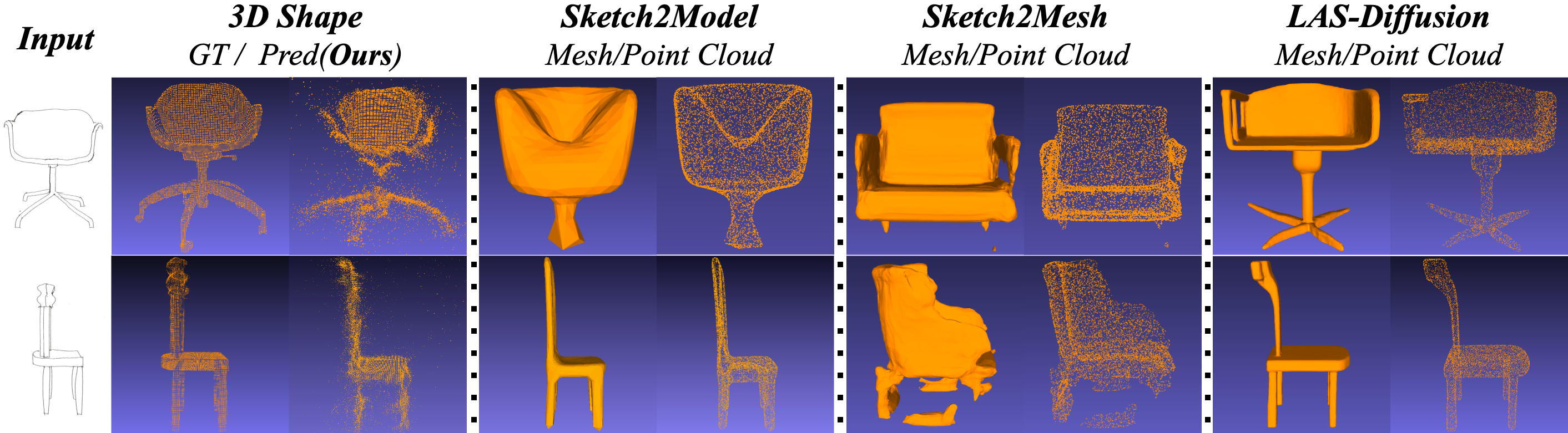}
	\caption{Reconstruction results on hand-drawn chair sketches \cite{zhong2020towards}.}
    \label{FIG_7}
\end{figure}

\begin{figure}[!t]\centering
	\includegraphics[width=12cm]{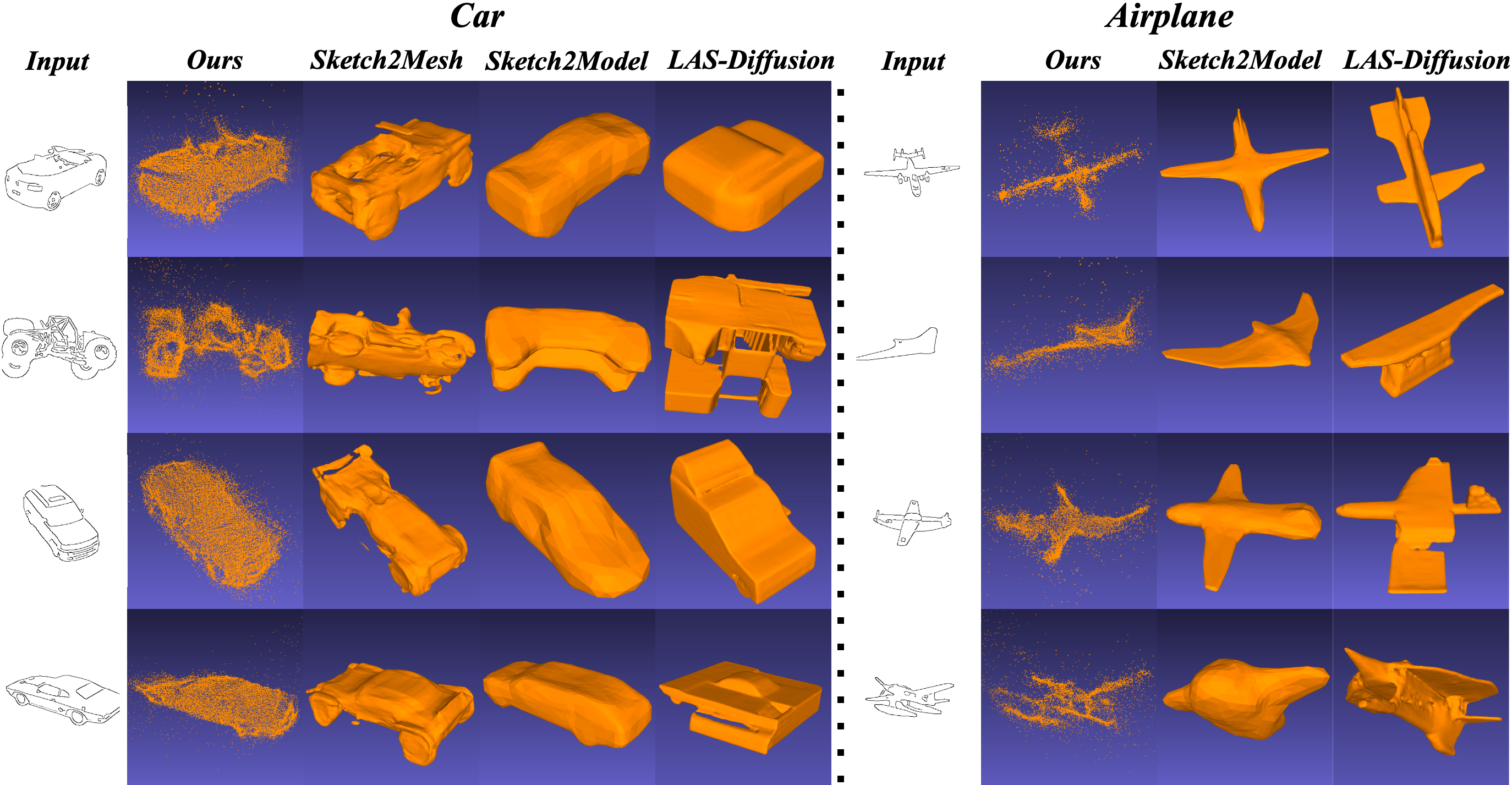}
	\caption{More qualitative results of synthetic car and airplane sketches.}
    \label{FIG_8}
\end{figure}

\noindent \textbf{Qualitative Comparison.}
Figure \ref{FIG_6} and Figure \ref{FIG_7} show the qualitative comparison on the chair category. Our method outperforms all baselines in shape recovery and geometry preservation. Particularly for detailed sketches with complex shapes, DiffS-NOCS captures geometric features more accurately. This is because our model fully utilizes the spatial alignment of input sketch and output point cloud, via generated NOCS map. The NOCS maps generated from the sketches are shown in Figure \ref{FIG_5}. And more qualitative comparisons of car and airplane categories are provided in Figure \ref{FIG_8}.

\subsection{Ablation Study and Analysis}

\begin{table}[t!]
  \centering
  \caption{Ablation study on chair category. The table illustrates the difference in model performance after removing sketch viewpoint encoder module. SVE denotes the Sketch Viewpoint Encoder.}
  \begin{tabular}{m{3cm}<{\centering}|m{1.5cm}<{\centering} m{1.5cm}<{\centering} m{1.5cm}<{\centering}}
    \hline
        Methods & CD $\downarrow$ & EMD $\downarrow$ & 2D IoU $\uparrow$\\
        \hline
        w/o SVE  & 2.677 & 9.64 & 0.804 \\
        \hline
        Ours  & \textbf{2.304} & \textbf{8.96} & \textbf{0.832} \\
        \hline
  \end{tabular}
  \label{table3}
\end{table}

\begin{table}[t!]
  \centering
  \caption{Ablation study on multi-view decoder.}
  \begin{tabular}{m{1.5cm}<{\centering}|m{3.2cm}<{\centering}|m{1.5cm}<{\centering} m{1.5cm}<{\centering} m{1.5cm}<{\centering}}
    \hline
    Category & Model & 2 views & 3 views & 5 views \\
    \hline
    \multirow{2}{*}{Chair} & Single-View($D_{LDM}$) & 2.127 & 1.972 & 1.924 \\
    & Multi-View($D_{MV}$) & \textbf{1.915} & \textbf{1.657} & \textbf{1.324} \\
    \hline
    \multirow{2}{*}{Car} & Single-View($D_{LDM}$) & 2.009 & 1.782 & \textbf{1.653} \\
    & Multi-View($D_{MV}$)  & \textbf{1.958} & \textbf{1.644} & 1.732 \\
    \hline
    \multirow{2}{*}{Airplane} & Single-View($D_{LDM}$) & 1.157 & \textbf{1.019} & 0.981 \\
    & Multi-View($D_{MV}$)  & \textbf{1.080} & 1.149 & \textbf{0.674} \\
    \hline
  \end{tabular}
  \label{table4}
\end{table}

\noindent \textbf{Design of Sketch Viewpoint Encoder.} Here we examine the effectiveness of the design of the sketch viewpoint encoder. The results are shown in Table \ref{table3}. It can be seen that SVE module is beneficial.

\noindent \textbf{Importance of Multi-view Decoder.} As shown in Table \ref{table4}, increasing the number of views significantly enhances reconstruction geometry. The multi-view decoder further improves the results by aggregating features with permutation equivariant layers, which effectively capture geometric relationships across different viewpoints. This highlights the critical role of multi-view decoder in achieving higher accuracy and robustness in the cross-modal 3D reconstruction task.

\section{Conclusion}
In this paper, we propose a diffusion-based method for sketch-to-point cloud reconstruction, converting sketches into NOCS maps with viewpoint awareness. Based on a ControlNet-inspired architecture, firstly, we train a viewpoint encoder to extract viewpoint features, which, combined with contour features from the Stable Diffusion Encoder, improve decoding performance. Then, a multi-view decoder facilitates cross-view information exchange, ensuring 3D consistency during generation. Our method excels in both single-view and multi-view reconstruction tasks, effectively handling hand-drawn sketches. Unlike 3D-based approaches, our method operates entirely in 2D using a controlled diffusion model, reducing both training and inference costs.

%
%
%
%

\end{document}